\documentclass[10pt, a4paper]{article}
\usepackage{mathtools}
\usepackage[final]{lrec2026} % this is the new style
\usepackage{xcolor}
\usepackage{booktabs}
\usepackage{float}
\usepackage{wasysym}
\usepackage{multirow}
\usepackage{amsmath}

\usepackage{fontspec}

\usepackage{polyglossia}
\setmainlanguage{english}
\setotherlanguage{arabic}

\usepackage{bidi}

\newfontfamily\arabicfont[Script=Arabic]{Amiri-Regular.ttf}[
  Path = fonts/,
  BoldFont = Amiri-Bold.ttf,
  ItalicFont = Amiri-Italic.ttf,
  BoldItalicFont = Amiri-BoldItalic.ttf
]

\usepackage{bidi}            % For RTL text in tables, etc.
\usepackage{polyglossia}     % Better multilingual support
\setmainlanguage{english}
\setotherlanguage{arabic}

\title{SLURP-TN : Resource for Tunisian Dialect Spoken Language Understanding}

\name{Haroun Elleuch$^{1,2}$, Salima Mdhaffar$^{2}$, Yannick Estève$^{2}$, Fethi Bougares$^{1,2}$}
\address{$^{1}$ ELYADATA, Paris, France, \\
         $^{2}$Laboratoire Informatique d’Avignon, Avignon Universite, Avignon, France \\
         \{haroun.elleuch, fethi.bougares\}@elyadata.com\\
         \{salima.mdhaffar, yannick.esteve\}@univ-avignon.fr\\}

\abstract{
Spoken Language Understanding (SLU) aims to extract the semantic information from the speech utterance of user queries. It is a core component in a task-oriented dialogue system.
With the spectacular progress of deep neural network models and the evolution of pre-trained language models, SLU has obtained significant breakthroughs. However, only a few high-resource languages have taken advantage of this progress due to the absence of SLU resources. In this paper, we seek to mitigate this obstacle by introducing SLURP-TN. This dataset was created by recording 55 native speakers uttering sentences in Tunisian dialect, manually translated from six SLURP domains. The result is an SLU Tunisian dialect dataset that comprises 4165 sentences recorded into around 5 hours of acoustic material. We also develop a number of Automatic Speech Recognition and SLU models exploiting SLUTP-TN. The Dataset and baseline models are available at: \url{https://huggingface.co/datasets/Elyadata/SLURP-TN}.
\newline \Keywords{SLURP-TN, Tunisian Dialect, Spoken Language Understanding, Automatic Speech Recognition\\}}

\begin{document}

\maketitleabstract

\section{Introduction}

% Spoken Language Understanding (SLU) is a core component of goal-oriented dialogue systems. SLU is generally defined as the combination of two sub-tasks: intent detection and slot filling \citep{tur2011spoken}.

% \begin{enumerate}
%     \item \red{Introduce SLU: when did it emerge + definition (Tur and De Morri book).}
%     \item \red{Explain the importance of SLU systems as a technology}
%     \item \red{Give a history of SLU datasets: ATIS, ..., MEDIA \& PORTMEDIA, SLURP, MASSIVE \& SpeechMASSIVE, TARIC-SLU (for low-resource Tunisian), Timers and Such}
%     \item \red{Explain limitations of those datasets: Restricted by domain, restricted to English / French / Italian (high resource), TARIC-SLU: restricted in domain (train tickets), recording quality, and speaker accent and gender distribution}
%     \item \red{Introduce SLURP-TN: A Tunisian multi-speaker dataset with 1. gender balance, 2. multiple sub-dialects, 3. acoustic conditions, 4. High sampling rate for other applications such as TTS + Meta-data available for tasks such as sub-dialect ID, gender ID, and Speaker ID. Moreover, the metadata availability enables finer-grained analysis by gender, age (although restricted), region, and scenario.} 
% \end{enumerate}

Spoken Language Understanding (SLU) is a fundamental component of goal-oriented dialogue systems, enabling machines to interpret user speech and extract structured meaning. Typically, SLU involves two sub-tasks: \emph{(i) intent detection}, which classifies the user’s communicative goal, and \emph{(ii) slot filling}, which identifies semantic arguments within an utterance~\citep{tur2011spoken}. Since its emergence in the early 1990s with task-oriented systems such as ATIS~\citep{hemphill-etal-1990-atis}, SLU has significantly evolved, driven by advances in automatic speech recognition (ASR), natural language understanding (NLU), and the growing availability of annotated datasets.

Several benchmark corpora have supported progress in this field, including ATIS for flight booking \cite{hemphill-etal-1990-atis}, MEDIA \cite{bonneaumaynard05_interspeech, laperriere2022spoken} and PORTMEDIA \cite{lefevre2012leveraging} for hotel reservation dialogues, and more recently SLURP \cite{bastianelli2020slurp}, Timers and Such \cite{lugosch2021timers}, and SpeechMASSIVE \cite{lee24i_interspeech}. Despite their contributions, these resources remain largely focused on high-resource languages such as English and French and are often limited to specific domains or recording setups. Low-resource languages and dialects, particularly Arabic varieties, remain underrepresented, which hinders the development of robust, inclusive SLU systems.

\citet{mdhaffar-etal-2024-taric} recently addressed this gap for Tunisian Arabic by introducing the TARIC-SLU corpus focused on the train ticket reservation domain. However, its narrow domain coverage, limited speaker diversity, and constrained acoustic conditions reduce its suitability for general-purpose SLU research and for tasks beyond intent and slot classification.

To address these limitations, we present SLURP-TN, a Tunisian Arabic multi-speaker corpus designed as a localized and translated extension of the SLURP dataset. SLURP-TN features balanced gender representation, coverage of multiple Tunisian sub-dialects, and recordings under diverse acoustic conditions (clean, noisy, and headphone). The corpus is sampled at a high frequency, making it suitable not only for SLU but also for other speech-related tasks such as text-to-speech and speaker identification. In addition, rich metadata annotations enable analyses across speaker gender, age, regional variety, and acoustic, providing a valuable resource for studying linguistic and acoustic variability in Tunisian Arabic. \\

Our contributions are threefold:
\begin{itemize}
    \item We create the first multi-domain SLU dataset for Tunisian Arabic dialects that can also be used for ASR, speaker identification and domain classification tasks; 
    \item We assess the performance of multiple SoTA systems under zero-shot and fine-tuned settings on SLU and ASR tasks;
    %provide baseline system evaluations on both the SLU and ASR tasks to provide a reference point for future research.
    \item We share the models and the dataset for academic use to further research on dialectal Arabic NLP.

\end{itemize}

% \newpage
\section{Similar Works} \label{sec:similar-works}

% While there have been efforts 
% To better situate our contribution, we compare our work with three relevant SLU corpora: SLURP, SpeechMASSIVE, and TARIC-SLU. \\
% \textbf{SLURP} is one of the most diverse SLU dataset to date in terms of speaker and domain diversity, with more than 18 scenarios. However, it is only in English.
% The text-only MASSIVE corpus \citep{fitzgerald2022massive} translates and localizes the SLURP corpus into 51 languages, one of the being Modern Standard Arabic (MSA).  \\
% \textbf{SpeechMASSIVE} is an audio recording effort meant to add speech coverage to the Massive dataset. It covers the entirety of validation and test splits of SLURP for 12 languages and completely covers French and German for the train set. The remaining 10 languages have 115 utterances recorded for the train split. This corpus does cover MSA.\\
% \textbf{TARIC-SLU} is the only known SLU corpus focusing on dialectal Arabic , namely Tunisian. While being larger than SpeechMASSIVE at 40.2 hours of train vs. less than half an hour in the latter, it focuses on only one domain: train reservations. 
% \\
%  Table~\ref{tab:slu_datasets_comparison} compares key characteristics ... 

To situate our contribution within the landscape of spoken language understanding (SLU) corpora, we compare relevant datasets:

\textbf{SLURP} is one of the most diverse SLU datasets to date, with over 18 distinct domains and a large number of speakers. However, it is available only in English, limiting its applicability to other languages \cite{bastianelli-etal-2020-slurp}. 

The \textbf{MASSIVE} corpus \citep{fitzgerald2022massive} extends SLURP by translating and localizing it into 51 languages, including Modern Standard Arabic (MSA). While MASSIVE provides extensive text-only coverage, the \textbf{SpeechMASSIVE} dataset adds audio recordings for multiple languages. For Arabic, SpeechMASSIVE covers MSA for all validation and test splits of SLURP, while the training set includes only 115 utterances, making it suitable primarily for few-shot experiments. 

\textbf{The JANA corpus}~\citep{jana} is a human-human dialogue corpus for Egyptian Arabic, consisting of 82 transcribed call center and chat dialogues annotated for dialogue. It does not, however, include code-switching nor semantic tag annotations.

\textbf{The \citet{10105079} dataset} is another dialectal Arabic dataset consisting of 10 Algerian Arabic voice commands recorded by 10 speakers, each repeated 10 times. Although it was evaluated on ASR, speaker identification, gender recognition, accent recognition, and SLU, its very small size limits its applicability for training robust task-oriented SLU models.

\textbf{The TuDiCol dataset} \cite{karoui:hal-01686469} is an ontology-annotated corpus for dialogues in Tunisian dialect. It comprises  127 dialogues for 893 segments. However, the dataset is not publicly available.

\textbf{TARIC-SLU} \cite{mdhaffar-etal-2024-taric} represents the only publicly available SLU corpus for Tunisian Arabic prior to our work. Although it contains a larger amount of audio data than SpeechMASSIVE (approximately 8 hours of training data versus 0.14 hours), it is limited to a single domain (train ticket reservations) and provides little information on speaker diversity. Its lower sampling rate (8 kHz) further constrains its use for broader speech applications.

\textbf{The PORTMEDIA dataset} \cite{lefevre2012leveraging} creation is similar to this undertaking, where the starting point is an existing dataset, MEDIA, recorded in another language, French, that is then translated to the Italian. However like the MEDIA and TARIC-SLU corpora, it focuses on a single domain, hotel reservations.

\begin{table}[htb]
\centering
\small
\begin{tabular}{@{}lccc@{}}
\toprule
\multicolumn{1}{c}{\textbf{}} &
  \multirow{2}{*}{\textbf{SLURP}} &
  \multirow{2}{*}{\textbf{\begin{tabular}[c]{@{}c@{}}Speech\\ MASSIVE (Ar)\end{tabular}}} &
  \multirow{2}{*}{\textbf{\begin{tabular}[c]{@{}c@{}}TARIC\\ SLU\end{tabular}}} \\
\multicolumn{1}{c}{} &         &          &            \\ \midrule
Arabic               & No      & MSA      & Tun.       \\
\#Domains            & 18      & 18       & 1          \\
\#Utt. (train)       & 11514   & 115      & 15751      \\
\#Speakers           & 177     & 61       & n/a        \\
Dur.[h] (train)      & 40.2    & 0.14     & 8          \\
% Recording Type     & HP + MP & xx       & \red{Street} \\
SR[kHz]              & 16      & 16       & 8          \\ \bottomrule
\end{tabular}
\caption{Comparison of the SLURP, SpeechMASSIVES's Arabic subset, and TARIC-SLU datasets.}
\label{tab:slu_datasets_comparison}
\end{table}

Table~\ref{tab:slu_datasets_comparison} summarizes key characteristics of the   datasets most relevant to our work (derived from SLURP or in Tunisian Arabic), highlighting the gaps that motivate the development of SLURP-TN.

\section{Corpus Creation}
In order to create our Tunisian dialect SLU dataset, we decided to benefit from the effort made during the creation of the SLURP dataset. This choice is motivated by the fact that SLURP is substantially bigger and more diverse than any other existing SLU dataset. Therefore, we used the same scenarios and the same set of semantic tags. 
% follow the annotation 
%Given the scale and diversity of the SLURP dataset~\citep{} (see Table~\ref{} \red{make a table for a comparison across datasets}), we adopted it as the foundation for the construction of our SLURP-TN corpus.
Due to practical constraints in time and resources, several design choices were made to ensure a balanced yet feasible data collection process:
\begin{itemize}
\item \textbf{Scenario selection.} We limited our coverage to six representative domains from the original 18 in SLURP—\textit{Emails, Weather, News, Books, Alarm}, and \textit{General}. These were selected for their relevance to everyday spoken interactions.
\item \textbf{Data volume per scenario.} For each selected domain, we translated and recorded up to 500 utterances, i.e., $\min(|\text{SLURP\_scenario}|, 500)$. This results in a maximum of approximately 500 utterances per scenario in SLURP-TN (see Figure~\ref{fig:bar_chart_scenarios}).
\item \textbf{Evaluation splits.} The entirety of the validation and test sets corresponding to the chosen scenarios was included in the recording process to allow direct comparison with the SLURP baselines.
\end{itemize}

\subsection{Participant Selection}
Participant selection is a critical step in the collection of speech datasets. Factors such as gender, regional origin, and linguistic proficiency are carefully considered to ensure that our dataset is of higher quality and representative of the target population. Overall, we assembled a team of 55 native Tunisian dialect speakers aged 23 to 35, and assigned them
the task of manually translating the source sentences into Tunisian dialect before proceeding to the recording sessions. 
%and achieves a wide demographic diversity coverage.
%suitable for their intended applications.

\subsection{Segment Translation}

Each speaker is responsible for translating their own utterances. Translations should reflect the speaker's natural way of speaking, including any instances of code-switching. Loanwords or code-switched segments that remain unchanged should be retained in their original language (English or French). During translation, participants are requested to adapt the sentences to the local Tunisian context. For instance, public figure names should be replaced by similar Tunisian figures, locations with Tunisian equivalents, and culturally specific references.
%—such as fast-food or restaurant chains— with locally relevant alternatives.

\subsection{Audio recording}
Audio recording sessions were scheduled according to the speaker's preferences. One speaker may participate in multiple recording sessions. All recording sessions are carried out under identical conditions. Audio is captured from the speakers through three different acoustic conditions: clean, noisy, and headphones. The clean audio condition refers to the audio signal captured using a professional Rhode VideoMic NTG microphone, while noisy and headphones refer to the audio signal recording, respectively, by the laptop and the headphones' microphones. The same laptop and headphones are used across recording sessions.

All the audio recordings are saved as a single-channel file in 24-bit Signed Integer PCM format with a sampling rate of 48khz. We used Reaper as a digital audio workstation and VoiceMeeter as a virtual soundcard to enable simultaneous recording from multiple inputs. We have deliberately chosen to use the same equipment and recording settings across sessions to eliminate any acoustically related bias in the data collection.
%microphone is performed collected under three acoustic conditions: noisy, , and clean. The ... \\
% \red{Recoding description, mic, time ... } \\
% \red{Sampling Rate + PCM format etc ...} \\
% \red{Audios recorded simultaneously etc ...} \\
% 24-bit Signed Integer PCM \\
% Mono \\
% 48 kHz \\
% \begin{itemize}
%     \item \textbf{Noisy (laptop or collar microphones):} Recordings include background noise, reverberation, and other environmental artifacts.
%     \item \textbf{Headphones:} Cleaner recordings with some presence of breathing sounds and occasional distortion.
%     \item \textbf{Clean (shotgun microphone):} Highest-quality recordings with minimal noise and distortion.
% \end{itemize}

\subsection{SLU annotation}
As stated earlier, we kept the same SLU tag set used for SLURP dataset annotation. As regards the SLU annotation step, it was performed after the completion of all recording sessions. 
This was realized by a single professional bilingual annotator, a native speaker of the Tunisian dialect. During this step, the annotator was provided with the annotated source utterances in English (as annotated in the original SLURP dataset) and their translation into the Tunisian dialect. They were asked to transfer the original English annotation to the correct position in the target sentences in the Tunisian dialect. Note that, compared to the source annotation, some slots could be dropped or change position in the Tunisian dialect sentence due to the re-ordering and lexical choices during English to Tunisian translation. Examples of Tunisian dialect sentences annotated with semantic tags alongside their original sentences in English are presented in Table~\ref{tab:example_utt}.
%\red{Sometimes the slots in the translated + localized sentence change due to the translation.} \\

\section{Corpus Statistics}

In total, we were able to collect 4165 Tunisian dialect sentences covering six different domains. Table~\ref{tab:slurptn-stats} presents details on the SLURP-TN dataset. A total duration of 2 hours and 46 minutes was recorded as a training set. This training set represents 2677 segments that were read out by 38 speakers. Among them, 20 are male speakers and 18 are female. On average SLURP-TN each training set recording is about 3.78 seconds in duration and contains about 5 words.

\begin{table}[h]
\centering
\begin{tabular}{lccc}
\hline
\textbf{Details} & \textbf{Train} & \textbf{Dev} & \textbf{Test}\\
\hline
Duration            & 2:46:33   & 44m       & 1:3:18                               \\
Speakers            & 38        & 6         & 11                                   \\
Gender \male/\female& 20 / 18   & 3 / 3     & 5 / 6                                \\ 
\# segments         & 2677      & 595       & 893                                  \\
Avg. dur. [sec]     & 3.78      & 4.43      & 4.25                                 \\
Avg. len. [wrd]     & 5.31      & 5.35      & 5.26                                 \\
% \\
\hline
\end{tabular}
\caption{SLURP-TN dataset statistics.}
\label{tab:slurptn-stats}
\end{table}

\begin{figure}[h]
  \centering
  \includegraphics[width=0.48\linewidth]{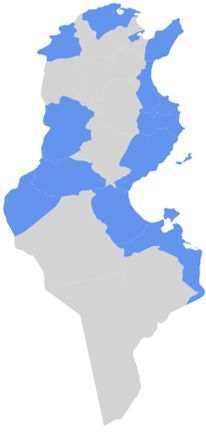}
  \caption{
  Geographic distribution of participants in the SLURP-TN recordings.}
  \label{fig:geo_cov}
\end{figure}

In addition to their gender and age differences, the speakers who participated in the SLURP-TN collection do not belong to the same geographic region in Tunisia. They are, in fact, spread across several Tunisian regions from north to south. This is an important feature that reflects the geographical representativeness and the coverage variability of our dataset. In total, we recorded speakers from 18 different regions that are shown in Figure \ref{fig:geo_cov}. %shows the geographic distribution of SLURP-TN speakers

\subsection{Domain coverage}

\begin{figure}[h]
    \centering
    \includegraphics[width=\linewidth]{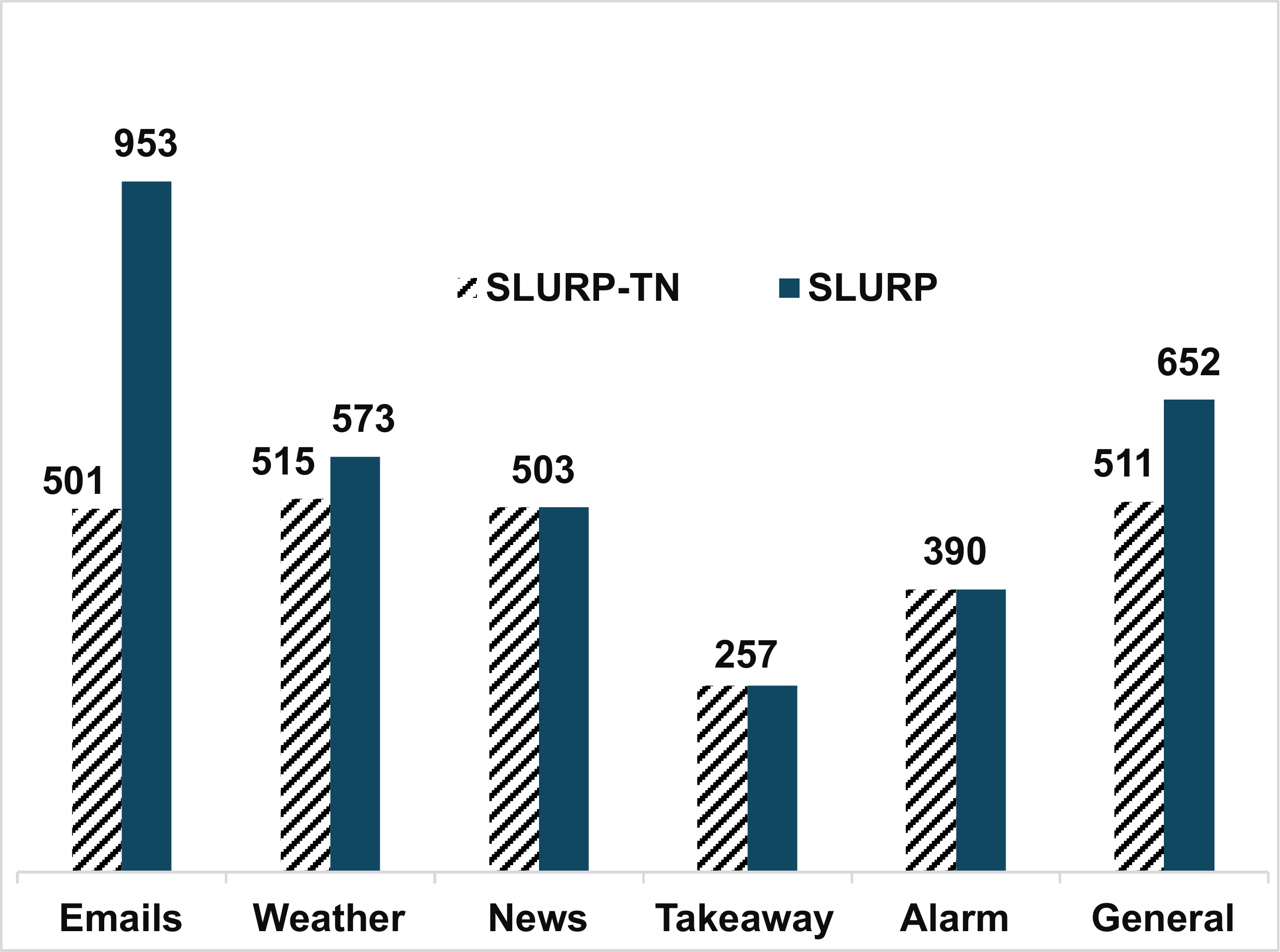}
    \caption{Train set recorded utterances by scenario in SLURP-TN vs. total number os sentences in the SLURP corpus.}
    \label{fig:bar_chart_scenarios}
\end{figure}

Among the 18 original SLURP domains, SLURP-TN covers 6 domains. Details about the selected domains are reported in Figure \ref{fig:bar_chart_scenarios}. As illustrated in the figure, we were able to fully translate and record all segments from News, Alarm, and Takeaway scenarios. For the three other domains, we achieve the goal of recording a least 500 sentences for each domain. For instance, 515 out of 573 sentences are randomly selected and covered in Weather, 511 out of 652 in General, and 501 out of 953 in the Emails domain. 
 
\subsection{Gender distribution}

Gender bias is the preference or prejudice toward one gender over the other~\citep{gender}. Gender bias often arises in NLP systems due to statistical patterns in the training corpora. In this work, we tried to mitigate gender bias by design by considering this factor during participant selection. 

\begin{table}[htb]
\begin{tabular}{@{}lccc@{}}
\toprule
\textbf{Domain} & \textbf{Train} & \textbf{Validation} & \textbf{Test} \\
                &  \multicolumn{3}{c}{male (\%) / female(\%) }    \\
\midrule
Emails          & 41.5/ 58.5    & 57.3 / 42.7     & 57.2 / 42.8   \\
Weather         & 54.6 / 45.4    & 47.6 / 52.4     & 55.8 / 44.2   \\
News            & 87.3 / 12.7    & 46.3 / 53.7     & 64 / 54       \\
Takeaway        & 26.5 / 73.5    & 50 / 50         & 45.6 / 54.4   \\
Alarm           & 30 / 70        & 34.4 / 65.6     & 40.6 / 59.4   \\
General         & 46 / 54        & 45.1 / 54.9     & 43.4 / 56.6   \\ \midrule
\textbf{Total}  & 50.4 / 49.6    & 48.2 / 51.8     & 49.9 / 50.1   \\ \bottomrule
\end{tabular}
\caption{Gender distribution (percent of male / female) in the SLURP-TN corpus by subset and domain.}
\label{tab:gender-dist}
\end{table}

Table \ref{tab:gender-dist} provides the percentage of male and female voices inside each domain in the train, validation, and test sets. Although the distribution is imbalanced for some domains (i.e train set of the Takeaway domain), the total number of recorded audios shows gender-balanced participation in the SLURP-TN dataset. 

\subsection{Code-Switching}
Code-switching (CS) is defined as the alternation of languages in text or speech. CS is a common linguistic phenomenon in multiple spoken languages, such as the Tunisian dialect \citep{bougares2025tedxtn}. SLURP-TN is not unique in that respect, and CS exists in almost 54\% of the recorded training sentences. Using the the changes in the writing script (Latin or Arabic), it is possible to identify instances of CS. The validation and test sets contain, respectively, 53\% and 58\% of CS sentences.

\begin{table*}[ht]
\centering
\begin{tabular}{p{2.5cm}p{10cm}}

\textbf{Domain} & \textbf{Sentences} \\ 
\midrule

\multirow{2}{*}{Alarm} 
 &\textbf{SLURP - } [time : six am] alarm please \\ 
 &\textbf{SLURP-TN - } [time : \textarabic{الستة متاع الصباح}] alarme \textarabic{يعيشك} \\ \midrule

%\multirow{2}{*}{Takeaway} 
% & will the [business\_type : restaurant] [order\_type : deliver] the order \\ 
% & [business\_type : resto]\textarabic{ال} \textarabic{الماكلة} [order\_type : \textarabic{يجيب}] \textarabic{بش} \\ \midrule
\multirow{2}{*}{Emails} 
 &\textbf{SLURP - } email [relation : co worker] about work project \\ 
 &\textbf{SLURP-TN - }  [relation : \textarabic{لزميلي}] \textarabic{لخدمة} projet \textarabic{على} email  \textarabic{ابعت} \\ \midrule

\multirow{2}{*}{General} 
 &\textbf{SLURP - } I'll be going to my [place\_name : office] \\ 
 &\textbf{SLURP-TN - } [place\_name : bureau] \textarabic{بش نمشي لل} \\ \midrule

%\multirow{2}{*}{News} 
% & what is the [media\_type : the times] headline \\ 
% & [media\_type : la presse] \textarabic{شنوا العنوان في} \\ \midrule

%\multirow{2}{*}{Weather} 
% & It's [weather\_descriptor : pleasant] [date : today] \\ 
% & [weather\_descriptor : \textarabic{تحفون}] [date : \textarabic{اليوم}] \\
\end{tabular}
\caption{Examples of SLURP and SLURP-TN sentences from General, Alarm and Emails domains .} % English sentences are on the first row, Arabic translations. } % for each domain}.%E}
\label{tab:example_utt}
\end{table*}

%%%%%%%%%%%%%%%%%%%%%%%%%%%%%%%%%
\begin{table*}[ht]
\centering
\begin{tabular}{@{}lcccccc@{}}
\toprule
  \textbf{Semantic tag}   & \textbf{Alarm} & \textbf{Takeaway} & \textbf{Emails} & \textbf{General} & \textbf{News} & \textbf{Weather} \\
  & \multicolumn{6}{c}{ train / dev / test}
  \\ \midrule
alarm-type         & 7/1/2     & --        & --        & --      & --        & --        \\
app-name           & --        & 4/0/0     & --        & --      & --        & --        \\
artist-name        & --        & --        & --        & 2/1/0   & --        & --        \\
business           & --        & 1/0/0     & --        & --      & --        & --        \\
business-name      & --        & 95/13/22  & 3/3/5     & --      & --        & --        \\
business-type      & --        & 37/5/2    & --        & 2/2/0   & --        & 0/1/0     \\
date & 102/16/26      & 2/0/0         & 41/16/19        & 72/13/32         & 59/11/15      & 250/71/74        \\
device-type        & 1/0/0     & --        & 0/1/0     & 2/0/0   & 1/0/0     & --        \\
drink-type         & --        & 1/0/1     & --        & 1/1/0   & --        & --        \\
email-address      & --        & --        & 19/3/9    & --      & --        & --        \\
email-folder       & --        & --        & 12/5/5    & --      & --        & --        \\
event-name         & 23/1/5    & 1/0/0     & 15/5/12   & 4/2/10  & --        & 3/0/1     \\
food-type          & --        & 107/15/20 & --        & 1/0/1   & --        & 0/1/0     \\
general-frequency  & 2/1/0     & --        & 1/0/0     & 0/0/1   & 2/0/1     & 1/0/1     \\
house-place        & 1/0/0     & --        & --        & --      & --        & --        \\
ingredient         & --        & 0/0/1     & --        & --      & --        & --        \\
joke-type          & --        & --        & --        & 27/8/11 & --        & --        \\
list-name          & --        & --        & 1/1/1     & --      & --        & --        \\
meal-type          & --        & 3/0/3     & 3/2/0     & 1/1/0   & --        & 1/0/0     \\
media-type         & 0/0/1     & --        & 1/0/0     & --      & 129/20/35 & --        \\
movie-name         & --        & --        & --        & 1/0/0   & --        & --        \\
news-topic         & --        & --        & --        & 1/0/2   & 173/36/45 & --        \\
order-name         & --        & 1/0/0     & --        & --      & --        & --        \\
order-type         & 1/0/0     & 97/15/16  & --        & --      & --        & --        \\
person             & 0/1/1     & 1/0/0     & 206/53/98 & 17/1/9  & 24/5/2    & --        \\
personal           & --        & --        & 1/0/0     & --      & --        & --        \\
personal-info      & --        & --        & 31/9/14   & --      & --        & --        \\
place-name         & --        & 1/1/0     & 3/1/3     & 14/8/6  & 65/8/13   & 115/22/30 \\
relation           & 2/0/0     & --        & 61/25/37  & 8/2/4   & --        & --        \\
time               & 170/32/41 & 6/0/0     & 25/6/9    & 2/1/1   & 8/1/2     & 14/3/3    \\
time-zone          & 1/0/0     & --        & --        & --      & --        & --        \\
timeofday          & 44/9/7    & 1/0/0     & 7/2/4     & 2/0/2   & 3/0/0     & 35/2/14   \\
transport-type     & --        & --        & --        & --      & 1/0/0     & --        \\
weather-descriptor & --        & --        & --        & --      & --        & 288/58/80 \\
\bottomrule
\end{tabular}
\caption{Distribution of the SLURP-TN semantic tags (train / validation / test).\\}
\label{tab:slu_tags}
\end{table*}

\section{Experimental Framework}

All our experiments are performed using a unified ASR and SLU experimental setup. All our training pipelines were implemented using the SpeechBrain toolkit~\citep{ravanelli2024open}. 
Depending on resource availability, models are trained on a single NVIDIA H100-80GB or A100-80GB GPU. 
%Model selection is guided by their strong performance reported in previous studies.
The SLURP-TN dataset was used to evaluate and fine-tune several SoTA multilingual speech models. Particularly, we evaluated w2v-BERT 2.0 \citep{barrault2023seamlessm4t} and SENSE  \citep{mdhaffar2025sense} models because they have been successfully used in previous related works. In fact, these models consistently outperformed other self-supervised (SSL) speech encoders on the TARIC-SLU dataset \citep{mdhaffar2024performance} and SpeechMASSIVE benchmarks \citep{mdhaffar2025sense}.\\
%. w2v-BERT 2.0 was also used to build SENSE~\citep{mdhaffar2025sense}—a semantically enriched SSL encoder and achieved superior results compared to state-of-the-art models on both the SLURP and SpeechMASSIVE benchmarks.

We additionally include Whisper-large-v3 due to its widespread adoption in the speech processing community. Whisper-based experiments are conducted by extending its tokenizer with SLURP-TN semantic tags or using only the encoder, as in the SENSE and w2v-BERT 2.0 configurations.\\

All our models are trained on the combined set of acoustic conditions (clean, noisy, and headset), using the corresponding combined validation set for model selection. 
Except for SENSE, all models undergo an initial fine-tuning stage in the TEDx-TN dataset~\citep{bougares2025tedxtn}, a Tunisian code-switching ASR and speech translation corpus. Previous work has shown that adapting generic models to data from the target language can significantly improve downstream performance~\citep{talafha2024casablanca}.

A character-level tokenizer is used for both ASR and SLU tasks. SLU-specific tags are encoded as single characters, resulting in total vocabulary sizes of 65 and 96 for ASR and SLU, respectively.

For encoder-based architectures, the pretrained encoder is followed by a downstream network consisting of three feed-forward layers with 512 neurons each and LeakyReLU activations, ending with a linear projection to the task’s vocabulary size and a log-softmax output. 
Encoders are optimized using Adam with a learning rate of $1 \times 10^{-5}$, while the downstream networks are trained using Adadelta with a learning rate of 1.0. 
Training is performed with the Connectionist Temporal Classification (CTC) loss.

In contrast, the encoder–decoder Whisper models are trained using the negative log-likelihood (NLL) loss and the AdaW optimizer.\\

All models are trained for 250 epochs with a batch size of 16 and a gradient accumulation factor of 2. 
Early stopping based on validation performance is applied to prevent overfitting. 
All audio signals are down-sampled to 16~kHz to ensure compatibility across models.

\vspace{0.3em}
\noindent
\textbf{Evaluation metrics.} 
ASR models are evaluated using the \textbf{character error rate (CER)} and \textbf{word error rate (WER)}. 
For SLU models, we report two complementary metrics: the \textbf{concept error rate (CoER)}, which measures the accuracy of detecting semantic tags within an utterance, and the \textbf{concept–value error rate (CVER)}, which evaluates the joint correctness of tag and value predictions.  In addition, WER and CER are computed for SLU models after removing semantic tags.
%The general formulation of an error rate is given by:
%\begin{equation}
%\text{ER} = \frac{S + D + I}{N}
%\end{equation}
%where $S$, $D$, and $I$ denote the number of substitutions, deletions, and insertions, respectively, and $N$ is the total number of reference units (characters, words, concepts, or concept–value pairs depending on the metric).
We use the SCLITE tool (version 2.4.10) provided by NIST~\cite{sctk} for all evaluations.

\section{Results}

\subsection{ASR Models}
\textbf{Zero-Shot condition:} We evaluate the w2v-BERT 2.0 system fine-tuned on TEDx-TN, an off-the-shelf Whisper-large-V3 model, and a fine-tuned version on TEDx-TD as well in a zero-shot fashion on our SLURP-TN dataset.\\

Table~\ref{tab:asr_zero_shot_results} reports the obtained results and shows that, despite its weakly-supervised training and support for the Arabic language, Whisper still lags behind on Arabic dialects such as Tunisian. The w2v-BERT 2.0 performed the best overall, with a relatively high WER score exceeding 59 in both evaluation splits. \\

\begin{table}[!htb]
\setlength{\tabcolsep}{6pt} % default is 6pt
\begin{tabular}{@{}lcccc@{}}
\toprule
\multicolumn{1}{c}{\multirow{2}{*}{\textbf{Model}}} & \multicolumn{2}{c}{\textbf{Validation}} & \multicolumn{2}{c}{\textbf{Test}} \\
\multicolumn{1}{c}{}                                & \textbf{CER}       & \textbf{WER}       & \textbf{CER}     & \textbf{WER}    \\ \midrule
w2v-BERT 2.0$^{*}$              & \textbf{15.7} & \textbf{59.2} & \textbf{17.4} & \textbf{59.5} \\
Whisp-lg-V3                     & 51.7 & 80.2 & 52.4 & 79.7 \\ 
Whisp-lg-V3 $^*$                & 32.8 & 67.3 & 32.4 & 67.3 \\
\bottomrule
\end{tabular}
\caption{Zero-shot WER (\%) and CER (\%) on the clean test and validation splits.}
\label{tab:asr_zero_shot_results}
\end{table}

Closer inspection of the system's output reveals that w2v-BERT 2.0 struggles with code-switched x, mixing Latin and Arabic script within the same word. \\
The off-the-shelf Whisper obtains the worst CER and WER scores overall, but closer inspection of the outputs shows a tendency to transcribe code-switched utterances entirely in the Arabic script. In other instances, dialectal Arabic sentences are transcribed in MSA. \\  
The Whisper system fine-tuned on TEDx-TN, achieves lower error rates than its off-the-shelf counterpart. It does, however, struggle with the pronunciation of code-switched segments, resulting in a large number of hallucinations. Despite the gains obtained by prior fine-tuning on the TEDx-TN dataset, Whisper systems still perform worse than w2v-BERT 2.0.

~\\
\textbf{Fine-tuning condition:} 
We fine-tune the models evaluated in a zero-shot fashion in addition to a SENSE system and Whisper-encoder only based system using the TEDx-TN whisper system.\\

\begin{table}[!htb]
\setlength{\tabcolsep}{6pt} % default is 6pt
\begin{tabular}{@{}lcccc@{}}
\toprule
\multicolumn{1}{c}{\multirow{2}{*}{\textbf{Model}}} & \multicolumn{2}{c}{\textbf{Validation}} & \multicolumn{2}{c}{\textbf{Test}} \\
\multicolumn{1}{c}{}                                & \textbf{CER}       & \textbf{WER}       & \textbf{CER}    & \textbf{WER}    \\ \midrule
%XLSR-128                        & 14.8 & 48.1 & 16.5 & 49.4 \\
%SAMU-XLSR                       & 15.3 & 48.8 & 16.8 & 49.6 \\
%w2v-BERT 2.0                    & 10.6 & 35.3 & 11.8 & 37.3 \\
w2v-BERT 2.0$^{*}$              & \textbf{8.4}  & \textbf{30.4} & \textbf{9.8}  & \textbf{33.6} \\
SENSE                           & 10.7 & 36.4 & 12.2 & 38.4 \\
%Whisp-lg-V3 enc.                 & 14.0 & 44.8 & 15.6 & 47.4 \\
Whisp-lg-V3 enc$^{*}$           & 13.4 & 44.0 & 15.4 & 45.8 \\
%Whisp-lg-V3                      & \red{wip}  &      &      &      \\
Whisp-lg-V3$^{*}$               & 22.1  & 51.2 & 21.2 & 40.4 \\ \bottomrule
\end{tabular}
\caption{WER (\%) and CER (\%) of ASR models on the clean test and validation splits.}
\label{tab:asr_results}
\end{table}

Table~\ref{tab:asr_results} summarizes the obtained results and shows that w2v-BERT 2.0 is the best performing system overall, nearly halving its zero-shot error rates after fine-tuning on SLURP-TN. The 30.4 and 33.6 WER on the validation and test splits show that SLURP-TN is still a challenging data set for ASR task. This can be attributed to the large amount of code-switched sentences and the speaker variability.\\

Although high performing on other SLU datasets like SLURP and SpeechMASSIVE, SENSE lags behind w2v-BERT 2.0 in the ASR evaluation. This can be attributed to the semantic enhancement being a trade-off between the model's ability to capture phonetic information for better performance on more semantics-oriented tasks.\\

Further fine-tuning the TEDx-TN Whisper system on SLURP-TN drastically improves ASR performance with a decrease of 16.1 and 26.9 absolute points in WER for the validation and test sets, respectively. \\
Moreover, the encoder-only whisper system outperforms its encoder-decoder counterpart in both CER and WER when evaluated on the validation split and in CER on the test split, where the encoder-decoder model achieves a lower WER. This indicates that for low-resource languages and code-switched settings, encoder-only systems still offer competitive performance.

\begin{table*}[htb]
\setlength{\tabcolsep}{2.45pt}
\centering
\begin{tabular}{l|ccc|ccc|ccc}
   \multicolumn{1}{c}{} & \multicolumn{3}{c}{CLEAN} & \multicolumn{3}{c}{HEADPHONE} & \multicolumn{3}{c}{NOISY} \\
\textbf{Model} & \textbf{CER/ WER} & \textbf{CoER} & \textbf{CVER} & \textbf{CER/ WER} & \textbf{CoER} & \textbf{CVER} & \textbf{CER/ WER} & \textbf{CoER} & \textbf{CVER} \\ \midrule
w2v-BERT 2.0$^{*}$     & \textbf{8.4}  / \textbf{30.2} & 53.9 & 61.3 & \textbf{8.5} / \textbf{31.4}  & 53.9 & 62.3 & \textbf{10.5} / \textbf{35.0} & 54.5 & 65.3 \\
SENSE                  & 10.8 / 36.3 & \textbf{45.1} & \textbf{55.7} & 10.8 / 37.0 & \textbf{43.8} & \textbf{54.9} & 13.2 / 40.9 & \textbf{47.0} & \textbf{60.2} \\
Whisp-lg-V3 enc.$^{*}$ & 14.1 / 45.5 & 56.8 & 70.3 & 14.1 / 45.2 & 57.6 & 70.7 & 19.1 / 54.6 & 63.7 & 77.7\\
Whisp-lg-V3$^{*}$      & 28.5 / 46.5 & 46.1 & 78.9 & 28.7 / 46.4 & 46.3 & 78.3 & 30.8 / 49.5 & 50.3 & 82.6 \\
\midrule
Average & \ul{15.5} / \ul{39.6}  & 50.5 & \ul{66.6} & \ul{15.5} / 40.0 & \ul{50.4} & \ul{66.6}  & 18.4 / 45.0 & 53.9  & 71.5 \\
\bottomrule
\end{tabular}
\caption{SLURP-TN \textbf{validation split} evaluation scores: WER, CoER, and CVER. Results in bold highlight the best model scores, and underlined values are the lowest on average.}
\label{tab:dev-results}
\end{table*}

\begin{table*}[htb]
\setlength{\tabcolsep}{2.45pt}
\centering
\begin{tabular}{l|ccc|ccc|ccc}
   \multicolumn{1}{c}{} & \multicolumn{3}{c}{CLEAN} & \multicolumn{3}{c}{HEADPHONE} & \multicolumn{3}{c}{NOISY} \\
\textbf{Model} & \textbf{CER/ WER} & \textbf{CoER} & \textbf{CVER} & \textbf{CER/ WER} & \textbf{CoER} & \textbf{CVER} & \textbf{CER/ WER} & \textbf{CoER} & \textbf{CVER} \\ \midrule
w2v-BERT 2.0$^{*}$          & \textbf{9.9}  / \textbf{33.4} & 53.2 & 66.5 &  \textbf{9.7} / \textbf{33.5}  & 52.7 & 65.3 & \textbf{11.2} / \textbf{36.3} & 51.3 & 67.8 \\
SENSE                       & 12.1 / 37.6 & \textbf{44.1} & \textbf{59.0} & 12.0 / 37.4  & \textbf{45.3} & \textbf{59.1} & 13.9 / 41.9 & \textbf{46.6} & \textbf{62.1} \\
Whisper-lg-V3 Enc.$^{*}$    & 16.3 / 48.3 & 57.2 & 71.8 & 16.2 / 47.5  & 57.4 & 71.8 & 20.0 / 55.0 & 61.5 & 78.4 \\
Whisper-lg-V3$^{*}$         & 28.2 / 46.1 & 50.3 & 78.8 & 28.1 / 45.8  & 49.1 & 77.9 & 30.2 / 49.1 & 50.3 & 78.5 \\ 
\midrule
Average & 16.6 / 41.4 & 51.2  & 69.0  & \ul{16.5} / \ul{41.1}  & \ul{51.1} & \ul{68.5}  & 18.8 / 45.6 & 52.4 & 71.7 \\
\bottomrule
\end{tabular}
\caption{SLURP-TN \textbf{test split} evaluation scores: WER, CoER, and CVER. Results in bold highlight the best model scores, and underlined values are the lowest on average.}
\label{tab:test-results}
\end{table*}

\subsection{SLU Models}

We evaluate all SLU systems under the three acoustic conditions-\textit{clean}, \textit{headphone}, and \textit{noisy}-for both the validation and test splits. The detailed results are presented in Tables~\ref{tab:dev-results} and~\ref{tab:test-results}, respectively.\\
~\\
\textbf{Performance of Semantically Enriched Models:} Across all acoustic conditions and evaluation splits, the SENSE model consistently outperforms w2v-BERT 2.0 on SLU-oriented metrics (CoER and CVER). However, this trend reverses when considering transcription fidelity metrics (CER and WER), where w2v-BERT 2.0 achieves lower error rates. 
This contrast illustrates a recurring trade-off between \textit{linguistic accuracy} and \textit{semantic understanding}: enriching SSL speech encoders with semantic knowledge enhances meaning-related comprehension at the expense of transcription precision.\\
~\\
\textbf{Encoder vs.\ Encoder-Decoder Architectures:} Whisper encoder-only systems demonstrate strong performance on the CER metric, nearly halving error rates compared to their encoder-decoder counterparts. However, the encoder-decoder variants achieve lower WER scores overall, suggesting that Whisper’s encoder-only outputs exhibit a higher dispersion of character-level errors across words.
In terms of semantic understanding, encoder--decoder systems outperform Whisper encoder-only models on CoER (tag prediction), yet lag behind in CVER (tag-value prediction). This pattern highlights complementary strengths between both architectures.\\
~\\
\textbf{Influence of Acoustic Conditions:}
When comparing acoustic conditions, the \textit{clean} and \textit{headphone} setups yield similar levels of linguistic (CER, WER) and semantic (CoER, CVER) complexity. On the validation split, models tend to achieve their lowest scores in either of these two conditions, while for the test split, the \textit{headphone condition} consistently produces the best results. In both cases, differences between the two remain minor: within 0.5 absolute points on average.  
The \textit{noisy condition}, however, remains significantly more challenging, resulting in consistently higher error rates across all models. Among them, \textbf{w2v-BERT 2.0} shows the strongest robustness under noise, exhibiting the smallest performance degradation. In contrast, \textbf{Whisper-based systems} suffer larger drops, with the Whisper encoder model showing up to a 7-point increase in CVER when moving from the headphone to the noisy condition.\\
~\\
\textbf{Validation vs.\ Test Set Behavior:}
Comparing results across splits reveals that the \textit{validation set} generally yields higher error rates, while the \textit{test set} exhibits less degradation across acoustic conditions. This suggests that the noisy portion of the test data may be acoustically cleaner or more balanced than its validation counterpart.\\
Overall, the validation and test results remain closely aligned across all models and metrics. This consistency indicates that the \textit{validation set serves as a reliable proxy} for estimating test set performance during model development and hyperparameter tuning.

~\\
\textbf{Complexity of the SLU Task:}
Comparing the CER and WER obtained by the ASR and SLU models on the \textit{clean} validation and test sets using Tables~\ref{tab:asr_results},~\ref{tab:dev-results}, and~\ref{tab:test-results}, can provide an estimation of the complexity of adding the slot-filling task.\\
The w2v-BERT 2.0 and SENSE systems are the least affected by the added SLU task, showing very little difference in scores.\\
While the Whisper-encoder system achieves noticeably worse CER and WER scores on the SLU task, especially on the test set, the encoder-decoder counterpart achieves much higher CER scores on the SLU evaluation sets. On the other hand, its WER scores are reduced by 5.1 absolute points when evaluating on the SLU validation set and increase by 5.7 on the test set of the same task. This behaviour is unique to the encoder-decoder and can be attributed to the extension of its tokenizer and embedding table instead of using a bespoke tokenizer like the other models we evaluate.

\section{Conclusion}
In this paper, we presented the SLURP-TN dataset, a translation and localization of the SLURP dataset featuring 55 speakers with an emphasis on gender parity and geographical coverage of Tunisian regions. In addition to the SLU annotations ported from SLURP, we provide high sampling rate audios in three different acoustic conditions (clean, noisy, and headphone) to enable experiments and analyses on other tasks.
ASR and SLU benchmarking of SoTA models shows that SLURP-TN is a challenging dataset for both tasks, highlighting the need for more robust and better-performing speech models for dialectal Arabic speech.

By making SLURP-TN publicly available for academic use, we hope to foster collaboration and facilitate more reproducible and comparable studies in SLU for Arabic dialects in general and the Tunisian dialect in particular.

\section{Limitations}

While our dataset exhibits the widest dialectal diversity to date for Arabic SLU, covering multiple dialects, a broad range of scenarios, and a large number of speakers with rich metadata, it still represents only about one-third of the original SLURP corpus in terms of overall scenario coverage. This reduced scope may limit the generalization of models trained on our data to certain domains or intents not yet represented. As future work, we aim to extend the dataset with additional scenarios and utterances to achieve broader coverage and closer alignment with the full SLURP benchmark.

\section{Ethical Considerations}

All recordings used in this work were collected with informed consent from participants. Personal identifiers were removed to preserve speaker anonymity, and all metadata is limited to non-sensitive information such as speaker age range, gender, and dialect region. We made efforts to ensure balanced representation across genders and dialects to mitigate demographic bias. However, as with any speech dataset, imbalances may remain due to differences in speaker availability or recording conditions.

The dataset should not be used in contexts that could lead to discrimination, surveillance, or any other harmful application.

\section{Acknowledgements}
This work was partially funded by the ESPERANTO project.
The ESPERANTO project has received funding from the European Union’s Horizon 2020 (H2020) research and innovation program under the Marie Skłodowska-Curie grant agreement No 101007666. This work was granted access to the HPC resources of IDRIS under the allocations 
AD011015051R1,
A0181012551, 
and AD011012108R3 
made by GENCI.

\stepcounter{section}
\renewcommand{\refname}{\thesection\quad Bibliographical References}

\bibliographystyle{lrec2026-natbib}
\bibliography{lrec2026-example}

\end{document}